\documentclass[11pt,a4paper]{article}
\usepackage[hyperref]{acl2021}
\usepackage{times}
\usepackage{latexsym}

\usepackage{microtype}

\aclfinalcopy 

\usepackage[utf8]{inputenc} 
\usepackage[T1]{fontenc}    
\usepackage{hyperref}       
\usepackage{url}            
\usepackage{booktabs}       
\usepackage{amsfonts}       
\usepackage{nicefrac}       
\usepackage{microtype}      
\usepackage{lipsum}
\usepackage{graphicx}
\graphicspath{ {./images/} }

\usepackage{microtype}
\usepackage{times}
\usepackage{soul}
\usepackage{url}
\usepackage[utf8]{inputenc}
\usepackage{graphicx}
\usepackage{amsmath}
\usepackage{booktabs}
\usepackage{algorithm}
\usepackage{algorithmic}
\usepackage[shortlabels]{enumitem}
\usepackage[TABBOTCAP]{subfigure}
\usepackage{multirow}
\usepackage{diagbox}

\def\AM{{\mathcal A}}

\def\EM{{\mathcal E}}

\def\RM{{\mathcal R}}
\def\TM{{\mathcal T}}

\def\VM{{\mathcal V}}
\def\WM{{\mathcal W}}

\def\0{{\bf 0}}
\def\1{{\bf 1}}
\DeclareMathOperator*{\argmax}{argmax} 

\title{Zero-shot Label-Aware Event Trigger and Argument Classification}

\author{Hongming Zhang\thanks{\; This work was done when the first author was visiting
the University of Pennsylvania.}$\; ^{,1,2}$, Haoyu Wang$^2$, Dan Roth$^2$ \\
  Department of Computer Science and Engineering, HKUST \\
  Department of Computer and Information Science, UPenn \\
  \texttt{hzhangal@cse.ust.hk, \{why16gzl, danroth\}@seas.upenn.edu}}

\date{}

\begin{document}
\maketitle
\begin{abstract}
Identifying events and mapping them to a pre-defined taxonomy of event types has long been an important NLP problem. Most previous work has relied heavily on labor-intensive, domain-specific, annotation, ignoring the {\em semantic meaning} of the event types’ labels. Consequently, the learned models cannot effectively generalize to new label taxonomies and domains. We propose a zero-shot event extraction approach, which first identifies events with existing tools (e.g., SRL) and then maps them to a given taxonomy of event types in a zero-shot manner.
Specifically, we leverage label representations induced by pre-trained language models, and map identified events to the target types via representation similarity. To semantically type the events’ arguments, we further use the definition of the events (e.g., argument of type ``Victim'' appears as the argument of event of type ``Attack'') as global constraints to regularize the prediction. The proposed approach is shown to be very effective on the ACE-2005 dataset, which has 33 trigger and 22 argument types. Without using any annotation, we successfully map 83\% of the triggers and 54\% of the arguments to the semantic correct types, almost doubling the performance of previous zero-shot approaches\footnote{Our code and models will be available at \url{http://cogcomp.org/page/publication_view/942}.}. 
\end{abstract}

\section{Introduction}

\begin{figure}[t]
    \centering
    \includegraphics[width=\linewidth]{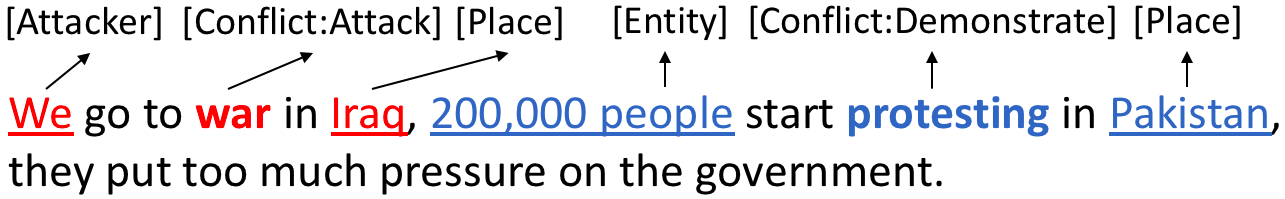}
    \caption{Event classification examples. Two events are highlighted with red and blue colors. Triggers and arguments are in bold and underline fonts, respectively.}
    \label{fig:intro_example}
\end{figure}

Event extraction, the process of identifying events triggers and arguments and classifying them into a set of pre-defined types is an important part of natural language understanding, and a commonly studied NLP task~\cite{grishman2005nyu}.
Consider the example shown in Figure~\ref{fig:intro_example}, where two events (i.e., ``war'' and ``protesting'') are identified.
By mapping them to [Conflict:Attack] and [Conflict:Demonstrate] and using the knowledge of ``Attack'' might result in ``Demonstrate'', we can infer that the war in Iraq is probably the cause of the protesting in Pakistan. 


Most existing event extraction work~\cite{DBLP:conf/emnlp/WaddenWLH19,DBLP:conf/acl/LinJHW20} 
treats event identification as a supervised sequence labeling task and event classification as a supervised classification problem, and relies on large amounts of event-specific annotated text.
Take ACE-2005~\cite{grishman2005nyu} as an example; the training set of ACE consists of 4,419 events, annotated and typed into 33 event types.
Such large-scale and high-quality annotation requires significant expertise, and it facilitates the success of supervised learning models.
However, scaling these efforts to new domains and more event types is very costly and unrealistic. 

Indeed, there has already been some effort to address the limitation of supervised models on new event types via a transfer-learning based zero-shot event classification approach~\cite{DBLP:conf/acl/DaganJVHCR18}.
By jointly encoding the event structures (i.e., the relations between event triggers and their arguments) of event mentions and of pre-defined event types, their model learns to map event mentions to the most similar event types.
As a result, at inference time, the model can be extended to new types as long as the structures of new types are provided.
Nonetheless, the success of this transfer learning approach also heavily relies on the similarity between observed event types and new ones.
When new event types are different enough from those the model was trained on, the model will struggle.

\begin{figure}
    \centering
    \includegraphics[width=0.75\linewidth]{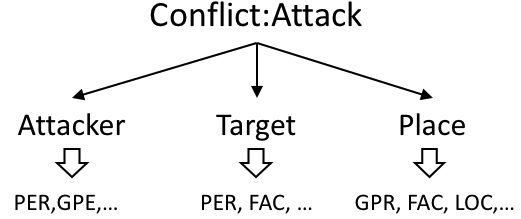}
    \caption{Pre-defined event type demonstration. Event type ``Conflict:Attack'' is associated with three argument types (``Attacker'', ``Target'', ``Place'') and each of them is associated with a list of potential entity types.}
    \label{fig:event_definition_example}
\end{figure}

This paper shows that the whole event extraction pipeline (i.e., identification and classification) can be done without any event-specific annotation. 
Since the event identification task can also be viewed as a classification task that determines whether the event triggers provided by the SRL models belong to the pre-defined event ontology or not, in this paper, we focus on the classification task.
Specifically, we explore a reliable zero-shot solution to mapping observed events to any given set of event types.
Unlike previous approaches, we do not use any annotation and only rely on the given event type {\em definitions}. We classify events by matching the semantics of the identified triggers and arguments to the type names, and then regularize the predictions with the constraints in the pre-defined event ontology.


A pre-defined event type example is shown in Figure \ref{fig:event_definition_example}, where domain experts choose to use ``attack'' to describe the whole event type and the labels ``attacker'', ``target'', and ``place'' for its roles, since the semantics of these words reflect ones’ understanding of this event type.
To fully utilize the semantics of these labels, we propose to represent the labels with a cluster of contextualized embeddings rather than just words.
In the aforementioned example, 
we first select several sentences that contain the word ``attack'' from an external corpus (e.g., New York Times (NYT)~\cite{sandhaus2008new}).
For each selected sentence, we use a pre-trained language model (e.g., BERT~\cite{DBLP:conf/naacl/DevlinCLT19}) to encode it and then use the resulting embedding of ``attack'' as a data point in the ``attack" cluster.
At the inference time, we can then acquire the contextualized representation of identified triggers and arguments, and map them to their corresponding types based on their similarities to those clusters in the embedding space.
Beyond the labels, event definitions also provide constraints between types and roles.
%
For example, the role ``Attacker'' can only appear as an argument of ``Conflict:Attack'' rather than ``Life:Marry,'' and only a person or a nation can take the role of an ``Attacker".
In our system, we propose to use these constraints to regularize the zero-shot model. Specifically, we formulate the final inference step as an integer linear programming (ILP)~\cite{RothYi04} and only produce decisions that satisfy all the constraints.

Our experiments show that the proposed model is very effective on the standard evaluation dataset ACE-2005, which has 33 event trigger types and 22 argument role types. Without using any annotation, we map 83\% of the triggers and 54\% of the arguments to correct types, almost doubling the performance of the previous best zero-shot event classification approach. When pipelined with our improved zero-shot identification step, we exhibit an zero-shot event extraction pipeline that rivals a SOTA supervised system~\cite{DBLP:conf/acl/LinJHW20} trained on over 6,000 sentences.  


\section{Task Definition and Notations}\label{sec:definition}

We denote the overall sets of predefined event trigger types, argument role types, and entity types as $\EM$, $\RM$, and $\TM$ respectively.
Each pre-defined event type (e.g., ``Conflict:Attack'') $E \in \EM$ is associated with several role types $R \in \RM_E$ and for each $R \in \RM$, the event definition also links it with several possible entity types $T \in \TM_R$.
Given a sentence $S$, which has a predicate $v$, several arguments $a$, and associated entity types $t$, the task of zero-shot event classification is mapping $v$ and $a$ to the correct types without any annotation.

\section{Model Overview}\label{sec:overview}

\begin{figure*}
    \centering
    \includegraphics[width=\linewidth]{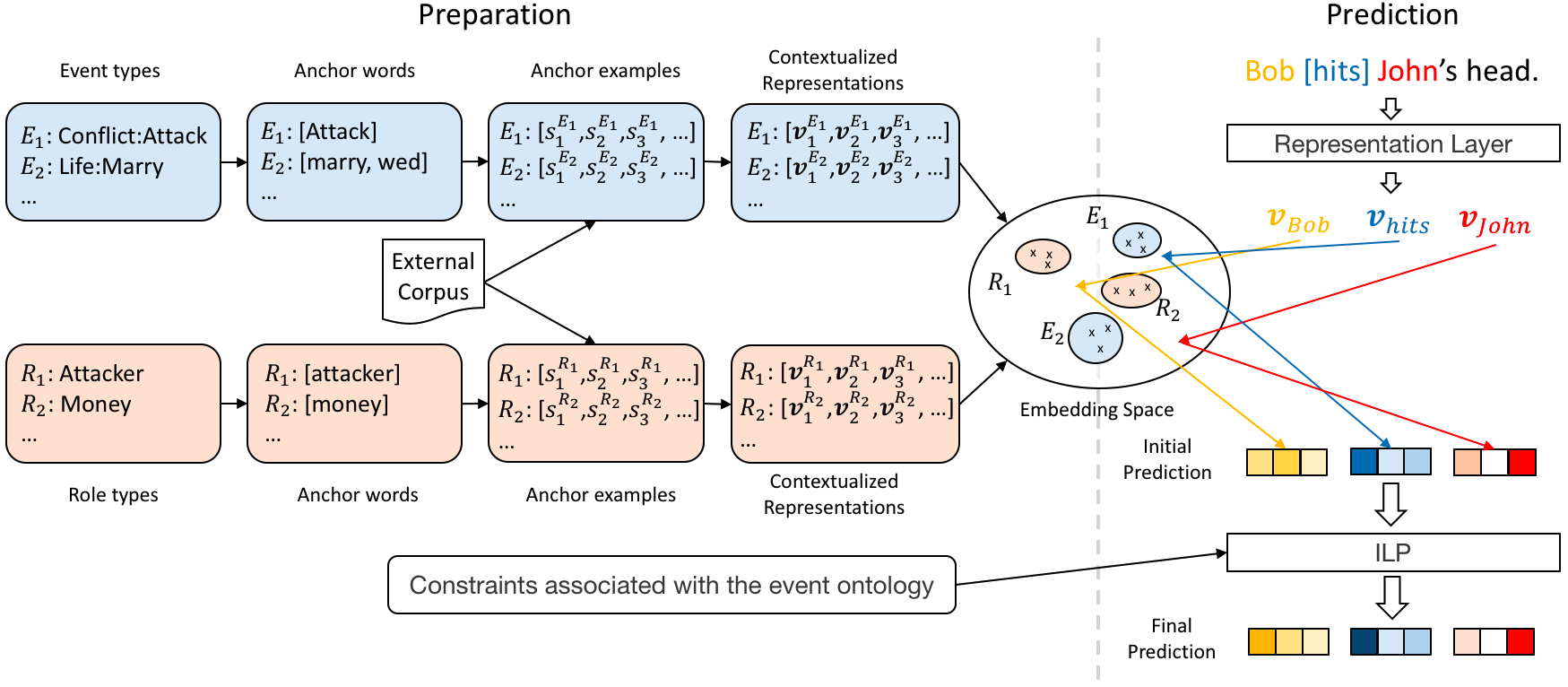}
    \caption{Architecture Overview. Pre-defined types, anchor words, anchor sentences, and contextualized representations for event trigger types and argument role types are indicated with blue and orange, respectively.}
    \label{fig:overall_framework}
\end{figure*}

As shown in Figure~\ref{fig:overall_framework}, the whole framework can be divided into two phases: preparation and prediction.
In the preparation phase, we generate the representations for all pre-defined event types and argument roles, which are indicated by blue and orange, respectively.
For each type, we select the label word and its synonyms\footnote{For labels that have multiple words, we select words that can best represent the label semantics.} as the anchor words, and then for each of them, we retrieve a list of anchor sentences that use these words from an external corpus.
After applying the pre-trained language models to encode all anchor sentences, we can then obtain the contextualized representations of selected anchor words and treat the cluster of these embeddings to as the type representation.
In the prediction phase, given $S$, $v$, and $a$, we first acquire the contextualized representation for the triggers and arguments. After that, in the embedding space, we can easily map them to the most similar types based on the cosine distance to each cluster.
Based on these initial predictions, we then leverage constraints provided by the event definitions to regularize the prediction by modeling it as an ILP problem.

\section{Preparation}\label{sec:label_representation}

In this section, we introduce the preparation details.

\subsection{Anchor Words and Sentences Selection}

Since the most frequently-occurred POS of triggers is verb and they typically have multiple senses, we also include their synonyms of the same meaning as anchor words to  improve the overall representation quality.
In total, we got 107 anchor words for 33 event trigger types and 22 anchor words for 22 event argument types.
After that, we then go to an external corpus (NYT corpus~\cite{sandhaus2008new} in our experiment) for finding anchor sentences that contain the corresponding anchor words.

\subsection{Contextualized Representation Generation}

\begin{figure}
    \centering
    \includegraphics[width=\linewidth]{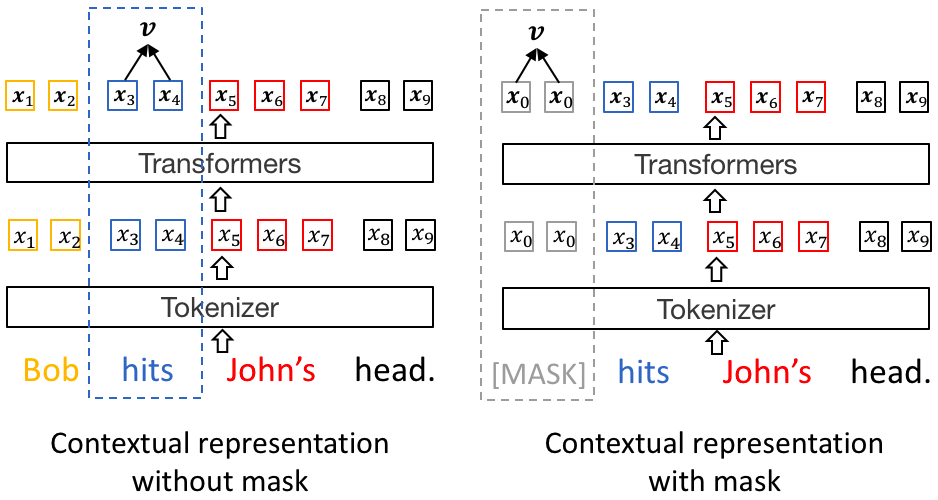}
    \caption{Demonstration of the proposed two contextualized representation acquisition methods.}
    \label{fig:representation_acquisition}
\end{figure}

As shown in Figure \ref{fig:representation_acquisition}, we propose two different representation acquisition methods for triggers and arguments. 
For triggers, we use all words in $S$ as the input, while for arguments, we mask the target anchor words.
The motivation is that most triggers contribute the most important semantic meaning while the semantics of arguments are often inferred from their context rather than the anchor words themselves.
For example, many arguments are pronouns or names, which only have weak semantics by themselves and we need to understand them by understanding their context.

For each sentence $S$ and a target anchor word $w^\star$, we first tokenize $S$ based on the type of the target word:
\begin{equation}
\small
    X, p_s, p_e=\left\{
\begin{aligned}
& Tokenize\_Full (S,w^\star) \ \mbox{for} \ w^\star \in \WM^T \\
& Tokenize\_Mask (S,w^\star) \ \mbox{for} \ w^\star \in \WM^A, 
\end{aligned}
\right.
\end{equation}

where $\WM^T$ and $\WM^A$ are sets of anchor words for triggers and arguments, respectively. $X$ = $x_1, x_2, \cdots, x_n$ is the list of tokens. As the tokenizer tool may tokenize words into sub-word pieces, we use $p_s$ and $p_e$ to record the start and end token positions for the anchor word. 
We then input $X$ into a multi-head transformer module to get the contextualized representations of all tokens. 
\begin{equation}
    \textbf{x}_1, \textbf{x}_2, \cdots, \textbf{x}_n = Transformer(x_1, x_2, \cdots, x_n).
\end{equation}
We omit the technical details of transformers for the clear representation, but the details are available in the original paper~\cite{DBLP:conf/nips/VaswaniSPUJGKP17}. 
In the end, we took the mean pooling of embeddings for tokens belonging to the target anchor word as one of its contextualized representation:
\begin{equation}
    \textbf{v} = \frac{\sum_{p_e  \leq i \le p_s} \textbf{x}_i}{p_e-p_s}.
\end{equation}
By grouping the acquired representations from all anchor sentences together, we get a cluster of embeddings for each pre-defined type, which can be used for the prediction. For trigger type $E_i$ and argument type $R_i$, we denote the corresponding vector cluster as $\VM^{E_i}$ and $\VM^{R_i}$, respectively.

\section{Prediction}\label{sec:ILP}

\begin{table}[t]
\small
    \centering
    \begin{tabular}{p{3.2cm}|p{4cm}}
    \toprule
       Explanation  & Constraint \\
         \midrule
        One type per trigger. & $\sum_{i \in |\EM|} I_t (i) = 1$\\
        \midrule
        One type per argument.& $\forall j \ \sum_{k \in |\RM| } I_r (j, k) = 1$\\
        \midrule
        Different arguments in one event must have different types. & $\forall k \ \sum_{j \in |\AM| } I_r (j, k) \leq 1$\\
        \midrule
        Predicted trigger and argument type must appear in the ontology .& $\forall i,j,k \ I_t(i) + I_r(j,k) \leq 1$ if $E_i$ and $R_k$ cannot be paired in the ontology.\\
        \midrule
        Entity types of arguments match the requirements. & $\forall i,j,k \ I_t(i) = 0, I_r(j,k) = 0$ if $a_i$ does not match the requirement of $E_k$ or $R_k$. \\
         \bottomrule
    \end{tabular}
    \caption{Selected constraints for the ILP regularization.}
    \label{tab:selected_constraints}
\end{table}

We then introduce the prediction part. 
For each identified event, whose trigger and $m$ arguments are denoted as $t$ and $\AM$ = $a_1, a_2, \cdots, a_m$, we first acquire the contextualized representations for trigger and arguments following the same way as what we did for anchor words.
Similar to the label representations, we acquire the event trigger representation without using masks and the argument embeddings with masks.
We denote the resulting embeddings as $\textbf{t}$ and $\textbf{a}_1, \textbf{a}_2, \cdots, \textbf{a}_m$.
After that, we compute the prediction score from $t$ to a pre-defined event trigger type $E$ as:
\begin{equation}
    f(t, E) = Cos\_Dist (\textbf{t}, \frac{\sum_{\textbf{v} \in \VM^E} \textbf{v}}{| \VM^E |}),
\end{equation}
where $Cos\_Dist$ represents the cosine distance and $| \VM^E |$ means the number of vectors in the cluster of $E$'s label representations.
Similarly, for any argument $a$ and pre-defined argument type $R$, we compute the prediction score via:
\begin{equation}
    f(a, R) = Cos\_Dist (\textbf{a}, \frac{\sum_{\textbf{v} \in \VM^R} \textbf{v}}{| \VM^R |}).
\end{equation}

After getting the initial predictions with cosine similarities, the next step would be leveraging the constraints to regularize the prediction results. 
Specifically, we model this problem as an integer linear programming (ILP) problem~\cite{RothYi04}, which maximizes the following objective while satisfying the constraints in Table~\ref{tab:selected_constraints}:


\begin{equation}
\scriptsize
\argmax_{I_t, I_a} \sum_{j \in |\AM|}( \sum_{i \in |\EM|} f(t, E_i) \cdot I_t (i) \cdot \lambda +  \sum_{k \in |\RM|} f(a_j, R_k) \cdot I_a(j, k)). 
\end{equation}

Here, $\lambda$ is the hyper-parameter we use to balance the weight of trigger and argument predictions, and $I_t$ and $I_a$ record the final prediction for the trigger and arguments, respectively. $I_t$ is a vector of integer variables with length $|\EM|$, and $I_a$ is a matrix of integer variables with the size $|\AM| \times |\RM|$.

\section{Experiment Details}\label{sec:experiment}

\begin{table}[t]
\small
    \centering
    \begin{tabular}{l|ccc|c}
    \toprule
                  & Train & Dev & Test & Overall \\
    \midrule
        \# Sentences  & 19,244& 902& 676& 20,822\\
       \# Event triggers  & 4,419 & 468& 424& 5,311\\
       \# Event arguments  & 6,604 & 759& 689& 8,052\\
    \bottomrule
    \end{tabular}
    \caption{ACE-2005 statistics. }
    \label{tab:ACE-statiscs}
\end{table}


We follow~\cite{DBLP:conf/acl/LinJHW20} and use ACE-2005 (E$^+$) as the dataset.
In total, ACE-2005 contains 33 event types and 22 role types.
The original dataset provides the official training, development, and test splits. However, as the proposed model is zero-shot and we do not need any training data, we merge all of them together to be the test set, which is consistent with the setting in \cite{DBLP:conf/acl/DaganJVHCR18}.
Detailed statistics about ACE-2005 are presented in Table~\ref{tab:ACE-statiscs}.
Considering that \cite{DBLP:conf/acl/DaganJVHCR18} trains the model with the most frequent ten event types and tests on the other 23 types, we provide two evaluation settings: (A) Evaluation on the least 23 frequent event types and associated role types, which is consistent with the previous work; (B) Evaluation on all 33 event types, which is used to demonstrate the overall performance of our model on the whole dataset.
We treat the trigger and argument classification as two separate ranking problems, and follow the previous work~\cite{DBLP:conf/acl/DaganJVHCR18} to report Hit@1, Hit@3, and Hit@5.

\begin{table*}[t]
\small
    \centering
    \begin{tabular}{l|cc|ccc|ccc}
    \toprule
       \multirow{2}{*}{Model}  & \multicolumn{2}{c|}{\# Types}&\multicolumn{3}{c|}{Event Triggers} &\multicolumn{3}{c}{Event Arguments} \\
                                                                                    & Train & Test & Hit@1 & Hit@3 & Hit@5 & Hit@1 & Hit@3 & Hit@5\\
    \midrule
    Frequency  & 0 & 23 & 9.6 & 27.2 & 42.5 & 25.9 & 63.4 & 80.6\\
    WSD-embedding & 0 & 23& 1.7 & 13.0 & 22.8 & 2.4 & 2.8 & 2.8 \\
    \midrule
    Transfer Learning (A) & 1 & 23 & 4.0 & 23.8 & 32.5 & 1.3 & 3.4 & 3.6 \\
    Transfer Learning (B) & 3 & 23 & 7.0 & 12.5 & 36.8 & 3.5 & 6.0 & 6.3 \\
    Transfer Learning (C) & 5 & 23 & 20.1 & 34.7 & 46.5 & 9.6 & 14.7 & 15.7 \\
    Transfer Learning (D) & 10 & 23 & 33.5 & 51.4 & 68.3 & 14.7 & 26.5 & 27.7 \\
    \midrule
    Label Representation  & 0 & 23 & 79.6 (0.6) & 88.2 (1.3) & 92.5 (1.7) & 25.9 (2.2) & 63.2 (1.9) & 74.6 (2.0)\\
    Label Representation + ILP  & 0 & 23 & \textbf{80.5} (0.2) & \textbf{88.9} (0.3) & \textbf{93.2} (0.6) & \textbf{68.5} (0.9) & \textbf{94.2} (0.1) & \textbf{96.8} (0.4) \\
    \midrule
    \midrule
    Frequency  & 0 & 33 & 28.9 & 53.6 & 62.7 & 13.8 & 33.8 & 51.0\\
    \midrule
    Label Representation & 0 & 33 & 81.9 (0.5) & 92.6 (0.4) & 95.7 (0.2) & 17.1 (0.7) & 38.0 (0.4) & 49.5 (0.9) \\
    Label Representation + ILP  & 0 & 33 & \textbf{82.9} (0.5) & \textbf{93.1} (0.1) & \textbf{96.2} (0.1) & \textbf{53.6} (1.3) & \textbf{87.9} (0.4) & \textbf{92.4} (0.5) \\
    \bottomrule
    \end{tabular}
    \caption{Event trigger and argument classification results on ACE-2005. Best performing models are annotated with the bold font. Standard deviations are shown in brackets.}
    \label{tab:main_result}
\end{table*}



\subsection{Baseline Methods}
To the best of our knowledge, only two previous methods were used to solve event typing in a zero-shot manner, thus we compare with both of them:
\begin{enumerate}[leftmargin=*]
    \item \textbf{WSD-embedding}~\cite{DBLP:conf/acl/DaganJVHCR18}: The WSD baseline first obtains the sense of event mentions with a word sense disambiguation module~\cite{DBLP:conf/acl/ZhongN10} and then acquires the sense embedding with the skip-gram model~\cite{DBLP:journals/corr/abs-1301-3781}. During the inference, it can then map triggers and arguments to the candidate types with the pre-trained word sense embeddings. 
    \item \textbf{Transfer Learning}~\cite{DBLP:conf/acl/DaganJVHCR18}: The transfer-learning based zero-shot approach first learns to map the AMR parsing~\cite{DBLP:conf/naacl/WangXP15} result of events to a few observed event types and then apply the learned model to unseen event types. In the original paper, four experiment settings are provided, which are distinguished by the number of seen event types, and we consider all of them to be the baselines.
\end{enumerate}

Besides them, we also present the performance of the ``Frequency'' baseline, 
which predicts all triggers and arguments with the most frequent types.

\subsection{Implementation Details}

We implement our model with Huggingface~\cite{DBLP:journals/corr/abs-1910-03771} and use BERT-large~\cite{DBLP:conf/naacl/DevlinCLT19} as the pre-trained language model. 
For each anchor word, we randomly select ten anchor sentences from the NYT corpus~\cite{sandhaus2008new}. 
$\lambda$ is set to be 100. We implemented the ILP optimization with gurobi\footnote{https://www.gurobi.com/}. 
All other hyper-parameters are inherited from BERT.
We repeat the experiments five times and report the average performance as well as the standard deviations.
The effect of all hyper-parameters is carefully evaluated.

\section{Result Analysis}

\begin{figure*}[tb]
    \centering
	\subfigure[Effect of the anchor sentence quantity $n$.]{\label{fig:anchor_example}
		\includegraphics[width=0.4\linewidth]{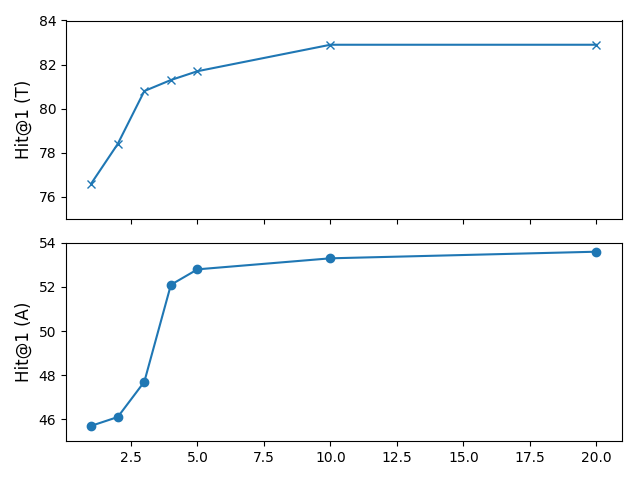}
	}        		
	\subfigure[Effect of the predicate weight $w$.]{\label{fig:weight}
		\includegraphics[width=0.4\linewidth]{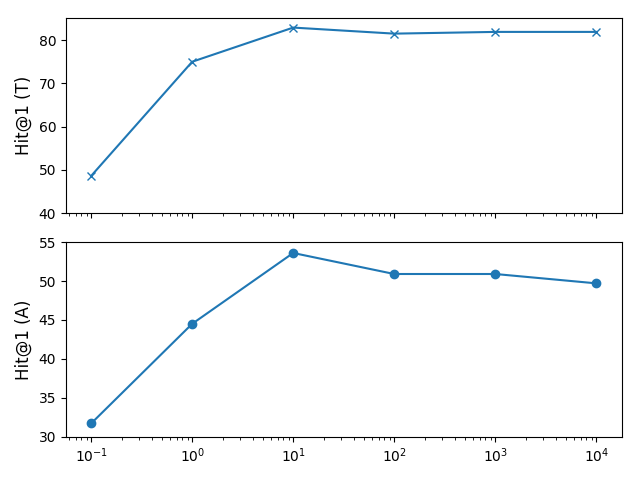}
	}        		  		
	\caption{Hyperparameter Analysis.} 
	\label{fig:hyperparameter}
\end{figure*}

From the results in Table~\ref{tab:main_result}, we can observe that:
\begin{enumerate}[leftmargin=*]
    \item Despite the difficulty of this task (we have 33 and 22 candidates types for triggers and arguments), our model can map 83\% of the triggers and 54\% of the arguments to the correct types, which shows that when the class labels are carefully designed, their semantics can serve as an excellent signal for the classification task.
    \item Compared with the baseline method, our model doubles the performance on the selected 23 event type subset. The main improvement we made is that we do not only use the labels but also put them back into some real usage and then use the contextualized representations to represent them. By doing so, we can best represent the label semantics and leverage them for the classification task.
    \item The dataset distribution is imbalanced. For example, 28.9\% of the event triggers are ``attack,'' which is relatively simple. This also explains why our model achieves higher performance for triggers on the whole dataset, where we need to map them to 33 types that include ``attack'', than the selected subset, where we only need to map them to 23 types but without ``attack''.
\end{enumerate}




\subsection{Ablation Study}

\begin{table}[t]
\small
    \centering
    \begin{tabular}{l||c|c||c|c}
    \toprule
        Model & Hit@1 (T) & $\Delta$ & Hit@1 (A) & $\Delta$\\
    \midrule
        Full model & 82.9 & - & 53.6 & - \\
        \midrule
        No Context & 40.5 & -42.4 & 29.9 & -23.7\\
        BERT-base  & 55.7 & -27.2& 33.5 & -20.1\\
        \bottomrule
    \end{tabular}
    \caption{Ablation study. The Hit@1 performance for triggers and arguments are denoted as Hit@1 (T) and Hit@1 (A).}
    \label{tab:context}
\end{table}

We present the following ablation studies to show the contribution of different modules:
\begin{enumerate}[leftmargin=*]
    \item \textbf{No Context}: One of the largest contributions of the proposed model is using contextualized representations to represent each label.
    To demonstrate the importance of the context, we try to remove them and acquire the label representation only with selected anchor words\footnote{Context are also removed for candidate triggers and arguments. For arguments, as there is no context, we also remove the masks.}.
    \item \textbf{BERT-base}: To demonstrate the contribution of a good representation model, we replace BERT-large with its weaker version (i.e., BERT-base~\cite{DBLP:conf/naacl/DevlinCLT19}).
\end{enumerate}

From the results in Table~\ref{tab:context}, we can see that the context information is crucial to our success. Without the context, the anchor word embeddings can no longer effectively represent the label semantics.
Besides that, a good language model also helps better merge the contextual information into the anchor words, which also shows that leveraging the context well is the key to our success.

\subsection{Hyper-parameter Analysis}

The effect of anchor sentence quantity $n$ and trigger weight $\lambda$ are shown in Figure~\ref{fig:anchor_example} and~\ref{fig:weight}, respectively.
First of all, we can observe that ten anchor sentences are enough to achieve a good performance,
which helps verify the motivation of this paper that with the careful usage, labels can serve as a crucial semantic signal for zero-shot classification tasks.
For $\lambda$, as shown in Table~\ref{tab:main_result}, before the ILP regularization, the model performs much better on triggers than arguments, that is why we need to give more weights to the trigger (i.e., the model should not change the trigger prediction unless it is very certain about the argument prediction).
On the other hand, $\lambda$ cannot be infinitely large, otherwise, the ILP may simply ignore the argument prediction, which may also hurt the performance.
To achieve the balance, we select $\lambda$ = 10. 



\begin{table}[t]
\small
    \centering
    \begin{tabular}{l|l||c|c}
    \toprule
        Trigger & Argument & Hit@1 (T)  & Hit@1 (A) \\
        \midrule
        w/o mask & w/ mask & 82.9  & 53.6 \\
        \midrule
        w/o mask & w/o mask  & 80.9 & 41.2 \\
        w/ mask & w/o mask &52.7 & 36.1 \\
        w/ mask & w/ mask & 52.8 & 40.0 \\
        \bottomrule
    \end{tabular}
    \caption{Effect of different representation strategies.}
    \label{tab:representation_acquisition}
\end{table}






\subsection{Representation Acquisition Strategy}

As aforementioned, we adopt two different label representation strategies for triggers and arguments.
Specifically, we kept the anchor words in the sentence for triggers while masking them for arguments.
To clearly show the effectiveness of different strategies, we show the performance of different strategy combinations in Table~\ref{tab:representation_acquisition}. 
From the results, we can see that if we apply masks to the trigger representations or remove the mask from the argument representation pine, the performance decreases.
This observation verifies the assumption that event triggers are often textually similar to the trigger labels and thus keeping these words in the sentence can help to map them.
On the other hand, event arguments are often named entities, which are often very different from the labels (e.g., ``victim'') textually, and what helps determine their roles is the context surrounding them. As a result, we achieve better performance when we mask these arguments and only leverage the context to generate the representations.



\section{Zero-shot Event Extraction}

\begin{table}[t]
\footnotesize
    \centering
    \begin{tabular}{l||cc|c}
    \toprule
                               & P & R & F1 \\
    \midrule
    Trigger (I)    & 58.9 & 57.8 &  58.3     \\
    Trigger (I + C) & 54.1 & 53.1 & 53.6\\
   
    \midrule
    Trigger + Argument (I)     & 12.0 & 26.0 & 16.4     \\
    Trigger + Argument (I + C)  & 4.6 & 10.0 & 6.3\\
    \bottomrule
    \end{tabular}
    \caption{Zero-shot event extraction performance. ``I'' and ``C'' mean the identification and classification.}
    \label{tab:event_extraction}
\end{table}

Previous experiments have demonstrated that with the help of the label representations and the post-regularization, our zero-shot system can effectively map detected triggers and arguments to the correct types.
However, if the goal is to automatically extract events from the raw documents, we still need the support from other NLP modules to identify the event triggers and arguments first.
In this section, we present the performance of a zero-shot event extraction pipeline, which combines our classification model with other NLP tools. 

\subsection{Identification}


The first step is identifying event triggers and arguments from the raw sentences.
In our pipeline, we use a BERT-based SRL model~\cite{DBLP:journals/corr/abs-1904-05255} and a nominal SRL model\footnote{As no previous work has applied BERT to the nominal SRL task, we trained one BERT-based nominal SRL model by ourselves.} to detect verbal and nominal events.
A limitation of these models is that they adopt a different event ontology (i.e., FrameNet~\cite{baker1998berkeley}) from ACE-2005. As a result, they could miss some events or detect irrelevant events, which are not annotated as events under the ACE definition. 
To include more ACE-specified events, we include all nominal anchor words\footnote{Some event type labels are nominal (e.g., ``bankruptcy'') and some are verbal (e.g., ``execute''). For nominal labels, we directly use them; for verbal labels, we use their nominal form (e.g., ``execution'') if available. As the used verb SRL system has already detected almost all verbal triggers, there is no need to add verbal keywords.} in the sentences as triggers. 
To filter out events that are not covered by the ACE ontology, we introduce an additional filtering step, whose details are introduced in the next sub-section.
Last but not least, considering that the definition of arguments in SRL systems varies from ACE, we include a mention detection module from CogCompNLP~\cite{KSZRCSRRLDTRMFWYSGUANLR18} to further detect mentions as argument candidates. 

\subsection{Filtering and Classification}

For triggers, following the previous work~\cite{DBLP:conf/acl/DaganJVHCR18}, we first combine the 1,161 event types from FrameNet and 33 event types from the ACE ontology together. Duplicated event types are manually removed. As a result, we got 1,147 event types. 
For each detected event trigger, we try to map it to all 1,147 event types with the proposed zero-shot classification approach.
Considering that 1,147 event types may still be not enough for covering all event types, we need to leave room for other events in the embedding space.
To do so, for each event type, we automatically acquire a cluster radius by optimizing the F1 score over all anchor sentences (i.e., the distance from representations of anchor sentences for that type to the mean representation should be smaller than the radius while others should be larger than the radius).
If a trigger is mapped to one of the 33 types in ACE and its cosine distance to the mean embedding of that type is smaller than the corresponding cluster radius, it will be categorized into that type.
For arguments, we select all mentions that overlap with the ARG0 and ARG1 of selected triggers given by SRL systems to be candidate arguments.
All selected triggers and arguments are jointly classified with the proposed classification model.


\subsection{Result Analysis}

\begin{figure}
    \centering
    \includegraphics[width=\linewidth]{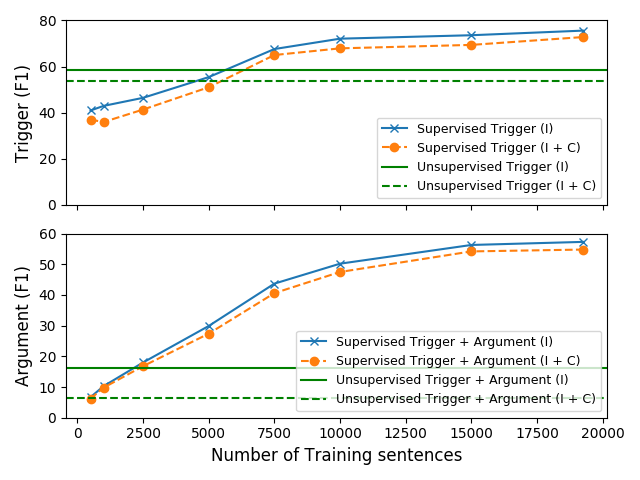}
    \caption{Comparison to the best supervised model. }
    \label{fig:supervised}
\end{figure}
\begin{figure}
    \centering
    \includegraphics[width=0.9\linewidth]{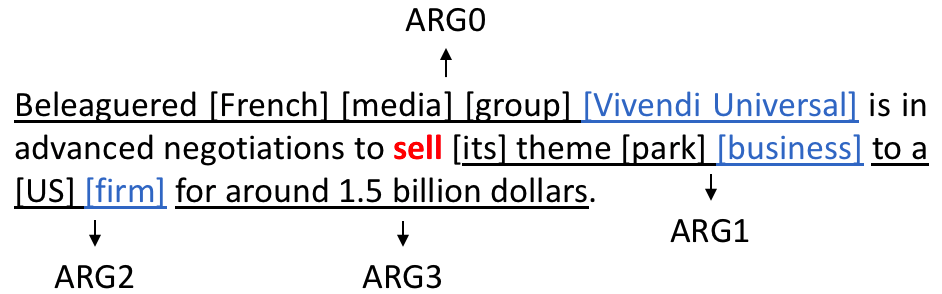}
    \caption{Case study for event extraction. Gold triggers and arguments are indicated with red and blue. Arguments detected by SRL and mention detection module are indicated with underlines and brackets.}
    \label{fig:case_study}
\end{figure}

The performance of our zero-shot event extraction pipeline is shown in Table~\ref{tab:event_extraction}.
``Identification'' requires the model to detect the correct spans of triggers and arguments.
``Classification'' requires the model to correctly classify the detected triggers and arguments.
As the identification task is often viewed as the sequence labeling problem, where the Recall@K metric is no longer suitable, we follow the previous work~\cite{DBLP:conf/acl/LinJHW20} to report Precision, Recall, and F1.
For the ``Argument'' evaluation, we provide gold triggers. But for the ``Trigger+Argument'' evaluation, the model has to detect and classify the correct trigger and associated arguments.
We also show the comparison of our system to the current best supervised model OneIE~\cite{DBLP:conf/acl/LinJHW20} in Figure~\ref{fig:supervised}.
For triggers, our system achieves the comparable performance of the supervised model that is trained with about 6,000 training sentences, which is equivalent to approximately 75\% of OneIE's full performance.
Considering that our system does not use any annotation and can be easily applied to new datasets and new event definitions, such performance is quite encouraging.
In the meantime, compared with triggers, our system cannot detect arguments very well, which is mainly due to poor identification performance.
As demonstrated in Figure~\ref{fig:case_study}, this is mainly because SRL and ACE adopt different definitions of arguments. SRL requires the arguments to cover all the details, whereas arguments in ACE are often just the key entities.
To solve this problem, we propose to use a mention detection module to detect mentions inside the arguments given by SRL systems.
Consequently, we cover more gold arguments but also introduce noise.
How to automatically identify arguments that fit the ACE definition is a problem worth exploring in the future.

\section{Related Works}\label{sec:related_works}

In this section, we introduce related works about the event extraction and NLP without annotation.

\subsection{Event Extraction}
Previous event extraction works often aim at learning supervised models, employing either symbolic features~\cite{DBLP:conf/acl/JiG08,DBLP:conf/acl/LiaoG10,DBLP:conf/acl/LiuCHL016} or distributed features~\cite{DBLP:conf/acl/ChenXLZ015,DBLP:conf/acl/LinJHW20}.
To address the problem that supervised models cannot be easily applied to new types, \cite{DBLP:conf/acl/DaganJVHCR18} separates the event extraction task into two parts (i.e., identification and classification) and proposes a zero-shot transfer-learning classification framework to apply the model trained with seen event types to unseen ones.
However, the prerequisite of their high performance is the similarity between seen and unseen event types.
Unlike previous works, we do not use any annotation and only leverage the label semantics to classify event triggers and arguments.
By combining our classification model and other NLP modules (i.e., SRL and mention detection), we achieve a decent zero-shot event extraction pipeline that can be easily applied to any new documents and event types.

\subsection{NLP without Annotation}

Solving NLP problems without using annotations has been explored in many NLP tasks including text classification~\cite{CRRS08,YinHaRo19}, entity typing~\cite{ZKTR18}, sequence classification~\cite{DBLP:conf/naacl/ReiS18}, and intent detection~\cite{DBLP:conf/emnlp/XiaZYCY18}.
The idea of leveraging the label semantics was first proposed in the dataless classification framework~\cite{CRRS08}, a predecessor name to what is now called zero-shot classification. The idea was to first map the text and labels into a common space using Explicit Semantic Analysis (ESA)~\cite{DBLP:conf/ijcai/GabrilovichM07} and then pick the label with the highest matching score.
This direction was later extended in~\cite{SongRo14,DBLP:conf/aaai/ChenXJC15,DBLP:conf/cikm/LiXSM16,DBLP:conf/coling/LiZTHIS16,SongMR16}. 
The most significant difference between our work and previous approaches is that, rather than using a fixed representation for each label, we use a group of contextualized embeddings as the representation.


\section{Conclusion}\label{sec:conclusion}

In this paper, we present a novel zero-shot classification model for event triggers and arguments.
By leveraging the rich semantics contained in labels and other constraints provided by the event definitions, we successfully classify 83\% of event triggers and 54\% of arguments to their correct types on the ACE-2005 dataset.
The ablation study demonstrates that the contextualized usage of the labels and correct way of using the context is key to our success.
Further experiments demonstrate that after combining the proposed zero-shot classification model with other available NLP tools, we can effectively extract and classify events without using any annotation. 
All the codes are submitted.

\section*{Acknowledgements}

This research is supported by the Office of the Director of National Intelligence (ODNI), Intelligence Advanced Research Projects Activity (IARPA), via IARPA Contract No. 2019-19051600006 under the BETTER Program, and by contract FA8750-19-2-1004 with the US Defense Advanced Research Projects Agency (DARPA). The views expressed are those of the authors and do not reflect the official policy or position of the Department of Defense or the U.S. Government.
We thank Celine Lee for providing the SRL model and the anonymous reviewers for their valuable feedback.


\bibliography{main,ccg}
\bibliographystyle{acl_natbib}  

\end{document}